\pdfoutput=1
\documentclass{article}

\usepackage{preprint,times}

\usepackage[export]{adjustbox}
\usepackage[ruled]{algorithm2e}
\usepackage[inline, shortlabels]{enumitem}
\usepackage[T1]{fontenc}
\usepackage{hyperref}
\usepackage{microtype}
\usepackage{pifont}
\usepackage{xcolor}
\usepackage{xurl}
\usepackage{graphicx}
\usepackage{booktabs}
\usepackage{float}
\usepackage{tabularray}
\usepackage{listings}
\usepackage{amsmath, amsfonts, amssymb}
\usepackage{nicefrac}

\UseTblrLibrary{booktabs}

\makeatletter
\newcommand{\ssymbol}[1]{\@fnsymbol{#1}}
\newcommand{\romanNumeral}[1]{\expandafter\@slowromancap\romannumeral #1@}
\makeatother
\usepackage{wrapfig}

\hypersetup{
    hidelinks,
    pdfauthor={Minbin Huang, Han Shi, Chuanyang Zheng, Yimeng Wu, Guoxuan Chen, Xingtong Yu, Yichun Yin, Hong Cheng},
    pdftitle={UniPool: Learning Expert-to-Layer Ownership from Brief Global Access}
}

\usepackage{physics}
\usepackage{mathtools}
\usepackage{subcaption}
\usepackage{multirow}
\usepackage{xspace}
\usepackage{wrapfig}
\newcommand{\method}{\textsc{UniPool}\xspace}
\newcommand{\methodfull}{\textsc{UniPool}\textsubscript{full}\xspace}
\newcommand{\methodlock}{\textsc{UniPool}\textsubscript{lock}\xspace}

\title{\method: Learning Expert-to-Layer Ownership\\from Brief Global Access}

\author{
    \textbf{Minbin Huang\textsuperscript{1}
    \quad Han Shi\textsuperscript{2}
    \quad Chuanyang Zheng\textsuperscript{1}
    \quad Yimeng Wu\textsuperscript{2}} \\
    \textbf{Guoxuan Chen\textsuperscript{3}
    \quad Xingtong Yu\textsuperscript{1}
    \quad Yichun Yin\textsuperscript{2}
    \quad Hong Cheng\textsuperscript{1}} \\
    \textsuperscript{1}The Chinese University of Hong Kong \\
    \textsuperscript{2}Huawei Technologies \\
    \textsuperscript{3}The University of Hong Kong
}

\begin{document}

\maketitle

\begin{abstract}
Most Mixture-of-Experts (MoE) transformers preassign each expert to one layer for the whole of training.
We study whether a brief global-access phase can learn a compatible expert-to-layer allocation that is then executed privately.
\methodlock gives every layer its own router over one global pool, balances aggregate pool usage, and scores experts with a scale-stable NormRouter.
After the short ownership-learning phase, it assigns each expert to one layer, locks the disjoint allocation, and continues training as a layer-private MoE.
With the same expert-FFN budget and routed expert FLOPs, this recipe lowers held-out loss relative to vanilla MoE by 0.024--0.037 across five dense-equivalent scales from 182M to 1.5B, including a 0.0247 reduction at 1.5B after 60B tokens.
The ownership-learning phase lasts 2K steps, about 3.3\% of training, and post-lock step time is within $-0.7\%$ to $+2.8\%$ of vanilla MoE in our throughput measurements.
Controls locate the gain in full-pool training and in the allocation it produces: a random disjoint allocation fixed at initialization, trained with the same router and losses, matches vanilla, whereas locking the allocation learned during the full-pool phase recovers nearly all of the persistent full-pool gain, and substituting a random allocation at the lock forfeits about half of it.
Keeping full-pool access throughout training (\methodfull) also improves over vanilla MoE at four scales and outperforms it with only 66.7\% (182M) to 50\% (469M and 650M) of its expert parameters.
The code is open-sourced at \url{https://github.com/Centaurus-Alpha/UniPool}.
\end{abstract}

\section{Introduction}
\label{sec:intro}

Mixture-of-Experts (MoE) models scale parameter capacity while keeping per-token computation sparse~\citep{shazeer2017outrageously,lepikhin2021gshard,fedus2022switch}.
Nearly all modern MoE transformers give each layer a private set of expert FFNs and a private router~\citep{jiang2024mixtral,dai2024deepseekmoe,deepseekai2024deepseekv2,deepseekai2024deepseekv3}.
This design commits each expert to one layer before any expert has been trained, so layer--expert ownership is never informed by data.

\begin{figure}[t]
    \centering
    \includegraphics[width=\linewidth]{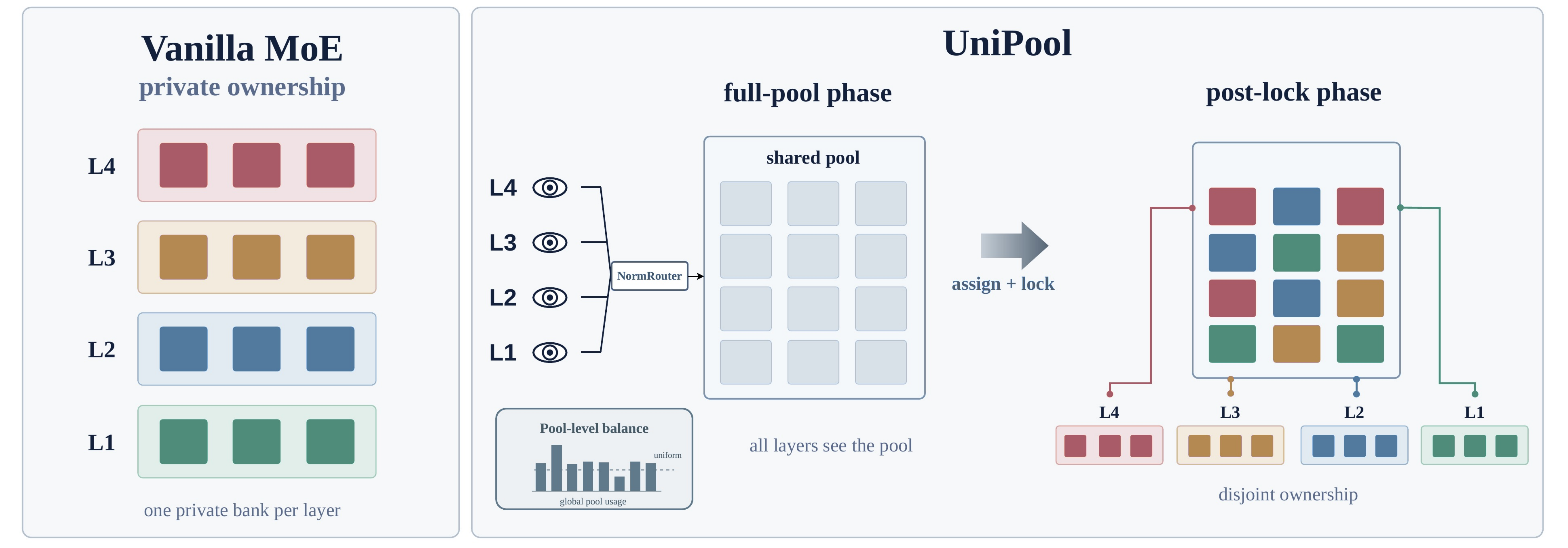}
    \vspace{-3mm}
    \caption{\textbf{Expert ownership in vanilla MoE and \method.} \emph{Left:} vanilla MoE restricts each layer to a private expert bank throughout training. \emph{Middle:} during a short full-pool phase, every layer routes over one shared pool through its own NormRouter, and a pool-level balance loss equalizes aggregate expert usage across depth rather than within each layer. \emph{Right:} \method then assigns each expert to exactly one layer from the observed routing statistics and locks this disjoint allocation (colors mark the owning layer); the remaining steps train a layer-private MoE. The persistent full-pool variant \methodfull keeps the middle-panel access pattern throughout training.}
    \label{fig:overall_framework}
    \vspace{-3mm}
\end{figure}

Several observations suggest that the fixed allocation wastes capacity.
Same-layer experts in trained MoEs share dominant weight subspaces, can be re-routed to similar experts with little loss, and can be pruned heavily, with redundancy concentrated in deep layers~\citep{huang2026sd,wu2026sere,bai2025diep}.
Our own routing probe (Section~\ref{sec:observation}) replaces the learned router of one deep layer in three production MoEs with uniform random routing and loses only 1.0--1.6 downstream accuracy points, so a deep layer's learned choice among its own experts adds little over a random one.
These findings describe a symptom.
We ask whether the cause is that ownership is fixed before training, when no layer can yet prefer any expert, and whether deciding it after the model has formed such preferences yields a better MoE at no lasting cost.

We propose \textbf{\method} (\textbf{Uni}fied Expert \textbf{Pool}; Figure~\ref{fig:overall_framework}).
Our main method, \methodlock, gives every layer its own router over one global pool during a brief ownership-learning phase, then locks a disjoint layer--expert allocation for the remaining training.
Two components make this phase trainable.
A \textbf{pool-level auxiliary loss} balances aggregate utilization across depth instead of forcing every layer to use every expert.
\textbf{NormRouter}~\citep{zheng2025understanding} computes L2-normalized, ReLU-sparsified scores whose scale does not depend on layer-specific logit norms.
Keeping every layer on the full pool throughout training is costly, however, because exposing all $M$ experts to every layer fragments grouped GEMMs.
\methodlock therefore uses full-pool access only as a short \textbf{ownership-learning phase}.
After it, \methodlock assigns each expert to exactly one layer from routing statistics, attenuates off-allocation scores, and locks a disjoint allocation.
From then on the model is a layer-private MoE with vanilla's number of experts per layer but an allocation that was learned rather than fixed.

The central empirical finding is that both the full-pool warm-up and the allocation it produces matter, and that their benefit survives when full-pool access is kept only briefly.
Freshly initialized experts are exchangeable, so any disjoint allocation fixed before training is vanilla MoE up to a relabeling of experts; accordingly, at 182M a random disjoint allocation fixed at initialization, trained with the same pool, router, and losses, matches vanilla MoE (1.9319 vs.\ 1.9317).
Locking the allocation selected at step 1K from full-pool routing statistics and enforced at step 2K improves loss to 1.9040, on par with persistent full-pool training (1.9029), whereas locking that same random allocation after the same 2K full-pool steps reaches only 1.9173.
A second seed of the same recipe reaches 1.9017, so the two seeds average to 1.9029, matching the persistent full-pool result.
\methodlock improves over vanilla MoE at all five scales we test, recovers nearly all of the persistent full-pool gain in every setting where both exist, and at 1.5B active parameters trained on 60B tokens it reduces test loss from 1.6320 to 1.6073.

Our contributions are as follows.
\begin{itemize}
    \item \textbf{Expert ownership as a learned quantity.}
    Standard MoEs commit each expert to a layer while the experts are still exchangeable, so the allocation carries no information; our matched controls show that deciding ownership after brief global access, from the routing preferences it produces, recovers the persistent full-pool gain, whereas substituting a random allocation at the lock forfeits about half of it.

    \item \textbf{\methodlock.}
    We combine a global expert pool, a pool-level balancing loss, and NormRouter routing with an ownership-learning phase and a disjoint lock, so that the model trains and executes as a layer-private MoE after the first 2K steps.

    \item \textbf{Consistent gains at vanilla-like cost.}
    With matched expert-FFN budgets and routed expert FLOPs, \methodlock improves over vanilla MoE from 182M to 1.5B dense-equivalent active scales, with post-lock step time within $-0.7\%$ to $+2.8\%$ of vanilla (Table~\ref{tab:throughput}).

    \item \textbf{Parameter-efficient persistent full-pool training.}
    Keeping full-pool access throughout training (\methodfull) improves over vanilla MoE at four scales and outperforms it with only 66.7\% (182M) to 50\% (469M and 650M) of its expert parameters, providing a parameter-efficiency reference for how expert budgets can scale with depth.
\end{itemize}

\section{Related Work}
\label{sec:related}
\paragraph{Sparse MoE and scaling.}
Sparsely gated expert layers~\citep{shazeer2017outrageously} were scaled through GShard, Switch Transformer, and ST-MoE~\citep{lepikhin2021gshard,fedus2022switch,zoph2022stmoe}, supported by systems such as DeepSpeed-MoE, Tutel, and MegaBlocks~\citep{rajbhandari2022deepspeedmoe,hwang2023tutel,gale2023megablocks}, and adopted by open models including Mixtral, the DeepSeek series, OpenMoE, and LLaMA-MoE~\citep{jiang2024mixtral,dai2024deepseekmoe,deepseekai2024deepseekv2,deepseekai2024deepseekv3,xue2024openmoe,zhu2024llama}.
Scaling studies examine routed capacity and expert granularity~\citep{clark2022unified,krajewski2024scaling,he2024mixture}.
These models commit experts to layers at initialization, before training can inform ownership; \method decides it after a brief global-access phase.

\paragraph{Routing and load balancing.}
The standard recipe combines softmax routing with the Switch auxiliary loss, which balances token fractions and routing probabilities within each layer~\citep{fedus2022switch}.
Alternatives include expert-choice routing~\citep{zhou2022mixtureofexperts}, linear assignment~\citep{lewis2021base}, hash routing~\citep{roller2021hash}, sigmoid gating with auxiliary-loss-free balancing~\citep{deepseekai2024deepseekv3,wang2024auxiliary}, and ReLU-based routing~\citep{wang2025remoe}.
When experts are shared across layers, dead-expert prevention should be defined over the whole pool rather than within each layer; \method therefore balances at the pool level and adopts NormRouter scores~\citep{zheng2025understanding}.

\paragraph{Cross-layer parameter sharing and expert reuse.}
Universal Transformers and ALBERT share entire blocks across depth~\citep{dehghani2019universal,lan2020albert}, and recursive variants relax such sharing with layer-wise adapters or token-level recursion depths~\citep{bae2025relaxed,bae2025mixture}.
MoEUT~\citep{csordas2024moeut} repeats a small group of shared MoE blocks across depth, and Mixture-of-Modules~\citep{gong2024mixture} assembles computation dynamically from a shared module set.
Closer to our setting, ReXMoE~\citep{tan2025rexmoe} lets each router also reach adjacent-layer experts, and MoUE~\citep{chen2026moue} combines layer-private experts with universal experts visible along structured depth windows.
DeepSeekMoE's shared experts are always active but remain within one layer~\citep{dai2024deepseekmoe}.
All of these methods prescribe the sharing pattern in advance.
\method instead learns an allocation from global competition and then removes sharing entirely, and Section~\ref{sec:ownership} compares it with predetermined patterns under a matched expert budget.

\paragraph{Changing MoE structure during training.}
Sparse upcycling initializes an MoE from a dense checkpoint~\citep{komatsuzaki2023sparse}, Drop-Upcycling partially re-initializes the copied experts to restore diversity~\citep{nakamura2025drop}, EvoMoE anneals a dense-to-sparse gate~\citep{nie2021evomoe}, and DS-MoE trains densely while inferring sparsely~\citep{pan2024dense}.
These methods change \emph{how many} experts a token uses over training.
\method keeps sparsity and expert compute fixed and changes \emph{which layer owns which expert}.

\section{Motivating Observation: Substitutable Experts Under Fixed Ownership}
\label{sec:observation}
\begin{wraptable}{r}{0.44\textwidth}
\vspace{-4mm}
\centering
\small
\setlength{\tabcolsep}{4pt}
\caption{Single-layer routing randomization in production MoEs: average raw zero-shot accuracy (\%) over five tasks. \textsc{Random} is averaged over deep-half layers; per-task results are in Appendix Table~\ref{tab:observation_production_full}.}
\label{tab:observation}
\begin{tabular}{@{}lccc@{}}
\toprule
\textbf{Model} & \textsc{Top-K} & \textsc{Random} & $\Delta$ \\
\midrule
Qwen1.5-MoE      & 67.92 & 66.29 & $-1.6$ \\
DeepSeek-V2-Lite & 54.19 & 53.03 & $-1.2$ \\
Qwen3-30B-A3B    & 73.02 & 72.06 & $-1.0$ \\
\bottomrule
\end{tabular}
\vspace{-3mm}
\end{wraptable}
Prior analyses find substantial within-layer redundancy in trained MoEs.
Same-layer expert weights in Qwen and DeepSeek MoEs share highly similar dominant spectral subspaces~\citep{huang2026sd}.
Re-routing tokens to the most similar same-layer expert preserves accuracy on Qwen1.5-MoE, DeepSeek-V2-Lite, and Qwen3-30B-A3B~\citep{wu2026sere}.
Pruning roughly half the experts of Mixtral~8$\times$7B costs about 8\% relative quality, with the strongest similarity in deep layers~\citep{bai2025diep}.
We complement these parameter- and output-level analyses with a probe of the router.
In Qwen1.5-MoE, DeepSeek-V2-Lite, and Qwen3-30B-A3B, we replace the learned top-$k$ router of one deep-half MoE layer with uniform random assignment, repeat this for every deep-half layer, and average downstream accuracy over layers (Table~\ref{tab:observation}).

The average drop is only $1.0$--$1.6$ points.
Under fixed private ownership, a deep layer's learned choice among its own experts thus adds little over a random choice: the experts it was given are only weakly differentiated.
This is consistent with the hypothesis that how experts are assigned to layers matters more than how many each layer holds.
If fixed ownership is the cause, learning ownership should make experts less substitutable even when every layer keeps the same number of experts.
Section~\ref{sec:ownership} tests the allocation rule under an identical expert budget, and Section~\ref{sec:routing_sensitivity} repeats this probe on our own models, including the locked model.

\section{Method}
\label{sec:method}

\method has three training components, global pool routing (Section~\ref{sec:pool}), pool-level balancing (Section~\ref{sec:pool_aux}), and NormRouter (Section~\ref{sec:norm_router}), plus an ownership lock (Section~\ref{sec:ownership_lock}).
The default method is \methodlock: it uses full-pool access only to learn an allocation and then locks that allocation for layer-private execution.
\methodfull keeps full-pool access throughout training; it serves as a quality reference and enables the reduced-budget analysis in Section~\ref{sec:full_pool_analysis}.

\subsection{Global Expert Pool}
\label{sec:pool}

In a standard MoE transformer with $L$ layers and $E$ experts per layer, layer $l$ owns experts $\{e_{l,1},\ldots,e_{l,E}\}$ and a router $r_l$:
\begin{equation}
    \text{FFN}_l(x) = \sum_{i \in \text{Top-}k(r_l(x))} g_{l,i}(x)\, e_{l,i}(x),
    \label{eq:standard_moe}
\end{equation}
where $g_{l,i}(x)$ is the gate that $r_l$ assigns to expert $i$.
\method replaces the $L$ private banks with one pool $\mathcal{E}=\{e_1,\ldots,e_M\}$ and keeps a separate router per layer:
\begin{equation}
    \text{FFN}_l(x) = \sum_{i \in \text{Top-}k_{\,\mathcal{V}_l}(r_l(x))} g_{l,i}(x)\, e_i(x),
    \label{eq:unipool}
\end{equation}
where $\mathcal{V}_l\subseteq[M]$ is the set of experts visible to layer $l$ and $\text{Top-}k_{\,\mathcal{V}_l}$ selects the $k$ highest-scoring experts within it.
During full-pool access $\mathcal{V}_l=[M]$ for every layer, so any layer may select any expert.
Routers stay layer-specific because hidden states differ across depth.
In matched comparisons $M=EL$, so \method stores the same expert FFNs and activates the same number per token as vanilla MoE; reduced-pool experiments set $M<EL$.

\subsection{Pool-Level Auxiliary Loss}
\label{sec:pool_aux}

The Switch auxiliary loss for layer $l$ is
\begin{equation}
    \mathcal{L}_{\text{aux}}^{(l)} = \alpha\, E \sum_{i=1}^{E} f_i^{(l)} P_i^{(l)},
    \label{eq:per_layer_aux}
\end{equation}
where $f_i^{(l)}$ is the fraction of layer-$l$ tokens dispatched to expert $i$ and $P_i^{(l)}$ is its mean routing score in that layer.
With private experts, balancing each layer is the right way to avoid dead parameters.
With a shared pool, an expert that one layer ignores may serve another, so per-layer balancing forces every layer to spread traffic over the whole pool and suppresses layer preferences.
We instead balance aggregate utilization.
With $\overline{f}_i=\frac1L\sum_l f_i^{(l)}$ and $\overline{P}_i=\frac1L\sum_l P_i^{(l)}$,
\begin{equation}
    \mathcal{L}_{\text{pool}} = \alpha_{\text{pool}}\, M \sum_{i=1}^{M} \overline{f}_i\,\overline{P}_i
    = \frac{1}{L} \sum_{l=1}^{L} \alpha_{\text{pool}}\, M \sum_{i=1}^{M} \overline{f}_i\, P_i^{(l)}.
    \label{eq:pool_aux}
\end{equation}
The decomposition lets each layer compute its own term.
Because $\overline{f}_i$ depends on all layers, we use its value from the previous micro-batch, which avoids cross-layer tensor dependencies under activation checkpointing.
Appendix~\ref{app:pool_aux_derivation} details this one-step-lagged estimate.

\subsection{NormRouter}
\label{sec:norm_router}

With softmax routing, layers with larger logit norms route more sharply, and in a shared pool these differences directly change which layer claims which experts.
We adopt NormRouter~\citep{zheng2025understanding}:
\begin{equation}
    s_i = \sigma\, c\, \max\!\left(0, \frac{z_i}{\|z\|_2 + \epsilon}\right),\qquad z=W_l h,
    \label{eq:norm_router}
\end{equation}
where $\sigma$ is a learnable scalar initialized to 1, $c$ is a fixed constant that sets the initial scale of the selected scores (Appendix~\ref{app:monte_carlo}), and $\epsilon$ ensures numerical stability. The norm $\lVert z\rVert_2$ is computed over all $M$ logits of layer $l$.
L2 normalization removes layer-dependent logit scale, and ReLU zeroes roughly half of the candidates.
Experts are selected by $s_i$, and both auxiliary losses use each expert's mean score in place of its mean softmax probability.
After the lock, we retain the $M$-row router and the same full-pool normalization, so the visibility mask changes the candidate set without changing score calibration.

\subsection{Learning Ownership and Locking It}
\label{sec:ownership_lock}
\label{sec:progressive_curriculum}

\paragraph{Why the default method locks the allocation.}
For a batch of $T$ tokens, top-$k$ routing over $M$ experts gives each expert about $Tk/M$ tokens.
With $M=EL$ visible experts per layer instead of $E$, the same routed FLOPs are split into $L$ times more, smaller grouped GEMMs.
Full-pool steps are 22\%--36\% slower than vanilla MoE (Table~\ref{tab:throughput}).
Masking unused experts does not help, because a conventional dispatcher still materializes $M$ columns and launches $M$ expert groups.

\paragraph{Ownership-learning phase.}
\method starts with full-pool access.
To make layer preferences concentrate before the lock, we add a target-cardinality regularizer
\begin{equation}
    \mathcal{L}_{\mathrm{card}}
    =\frac{\lambda_{\mathrm{card}}}{L}\sum_{l=1}^{L}
    \left(H(p_l)-\log K_{\mathrm{tar}}\right)^2,
    \label{eq:target_cardinality}
\end{equation}
where $p_l$ is layer $l$'s normalized routing-mass distribution over experts and $K_{\mathrm{tar}}=E$, so each layer's effective number of experts $\exp(H(p_l))$ approaches its eventual allocation size.
The pool loss simultaneously keeps aggregate usage balanced, which is exactly what a disjoint allocation needs: every layer concentrates on a few experts, and different layers concentrate on different ones.

\paragraph{Allocation, attenuation, and lock.}
At step 1K we collect routing counts on the held-out 3\% validation split (disjoint from the final test tokens) and solve a small assignment problem that gives every expert at most one owner layer.
Since $M=EL$ and each layer must receive at least $E$ experts, the solution gives every layer exactly $K_l=E$ experts and uses every expert once; among such allocations it maximizes the retained routing mass.
The allocation $\mathcal{C}_l$ is therefore disjoint across layers.
Over steps 1K--2K we attenuate off-allocation scores before top-$k$ selection,
\begin{equation}
    \widetilde{s}_{l,i}(t)=s_{l,i}(t)\left[1-\alpha(t)\,\mathbf{1}\{i\notin\mathcal{C}_l\}\right],
    \label{eq:curriculum_attenuation}
\end{equation}
with $\alpha(t)$ rising from 0 to 1 on a cosine schedule.
At step 2K we set $\mathcal{V}_l=\mathcal{C}_l$ permanently.
From then on each expert is trained and executed by one layer only, for the remaining 96.7\% of a 60K-step run.
The 1.5B run follows the same schedule.

\paragraph{Compact dispatch.}
After the lock, scores and the pool loss are still computed over the global $M$-expert index, with the mask applied before top-$k$ selection.
We gather the dispatch map into each layer's local coordinates, so token permutation, token-count metadata, and grouped GEMM operate on $|\mathcal{V}_l|=E$ expert groups that reference the original parameters without copying them.
Execution is then identical in shape to a vanilla MoE layer.
Appendix~\ref{app:progressive_curriculum_details} lists the schedule, assignment constraints, and implementation invariants.

\section{Experiments}
\label{sec:experimental}
\label{sec:results}

\subsection{Setup}
\label{sec:setup}

\paragraph{Models and data.}
We train LLaMA-style backbones~\citep{touvron2023llama} at five dense-equivalent active-parameter scales, 182M, 469M, 650M, 830M, and 1.5B (Appendix Table~\ref{tab:model_configs}).
The four scales up to 830M train on the Pile~\citep{gao2020pile} for 60K steps with global batch 512 and sequence length 1,024, about 30B tokens, using AdamW~\citep{loshchilov2019decoupled}, cosine learning-rate decay, and bf16 in Megatron-LM~\citep{shoeybi2020megatron}.
The 1.5B backbone has 36 layers of width 1,600 and trains with global batch 1,024 for 57,220 steps, about 60B tokens.

\paragraph{MoE configurations.}
Vanilla MoE uses eight private experts per layer, top-1 softmax routing, and a per-layer auxiliary loss.
\method uses a pool of $M=8L$ experts, top-1 NormRouter routing, and the pool loss.
Both store the same expert FFNs and activate one expert per layer and token; \method's wider routers add 0.10\%--0.37\% stored parameters (Appendix Table~\ref{tab:model_configs}).
We additionally add experts to the 12-layer 182M backbone with 16E/top-2 and 32E/top-4, keeping each expert at full FFN width; this raises vanilla MoE to about 268M and 438M active parameters, while \method has about 269M and 441M active and 1.46B and 2.82B total parameters, and each setting is compared only with vanilla MoE of the same configuration.

\paragraph{Seeds.}
The 182M vanilla and \methodfull results average three seeds, with test-loss standard deviation 0.0023.
The 182M \methodlock entry averages two same-recipe seeds (1.9017 and 1.9040); the other scale results use one seed, and control C in Section~\ref{sec:ownership} is the second seed of this recipe.
We treat differences below about 0.005 as ties.
All reported final losses are measured on the held-out test split, which is disjoint from the 3\% validation split used to choose allocations.

\subsection{Main Results}
\label{sec:main_results}

\begin{table}[t]
\centering
\caption{Test loss (lower is better). Within each column, the MoE variants match the expert-FFN budget and routed expert FLOPs; the wider shared-pool routers add 0.10--0.37\% stored and up to 2.2\% active parameters (Appendix Table~\ref{tab:model_configs}). \methodlock is the main method: it locks a learned disjoint allocation at step 2K; \methodfull is the persistent full-pool reference and was not trained at 1.5B. The 182M \methodlock entry is the mean of two same-recipe seeds (1.9017 and 1.9040). The five scales use the default 8E/top-1 budget, with 30B training tokens up to 830M and 60B at 1.5B; the last two columns add full-width experts to the 12-layer 182M backbone, giving about 268M (16E/top-2) and 438M (32E/top-4) active parameters. Dense models match the active parameters of the 8E/top-1 MoEs up to 830M. $\Delta$ is \methodlock minus vanilla MoE.}
\label{tab:main_results}
\vspace{4pt}
\small
\setlength{\tabcolsep}{4.5pt}
\begin{tabular}{@{}lccccc@{\hspace{14pt}}cc@{}}
\toprule
& \multicolumn{5}{c}{\textit{8E/top-1}} & \multicolumn{2}{c}{\textit{182M backbone}} \\
\cmidrule(lr){2-6}\cmidrule(l){7-8}
& \textbf{182M} & \textbf{469M} & \textbf{650M} & \textbf{830M} & \textbf{1.5B} & \textbf{16E/top-2} & \textbf{32E/top-4} \\
$L$/$H$ & 12/768 & 24/1024 & 36/1024 & 48/1024 & 36/1600 & 12/768 & 12/768 \\
\midrule
Dense & 2.0420 & 1.8860 & 1.8318 & 1.8032 & -- & -- & -- \\
Vanilla MoE & 1.9317 & 1.7982 & 1.7568 & 1.7309 & 1.6320 & 1.8589 & 1.7974 \\
\methodlock & \textbf{1.9029} & \textbf{1.7615} & 1.7324 & 1.6944 & \textbf{1.6073} & 1.8291 & 1.7702 \\
\methodfull & \textbf{1.9029} & 1.7636 & \textbf{1.7260} & \textbf{1.6923} & -- & \textbf{1.8277} & \textbf{1.7672} \\
\midrule
$\Delta$ & $-0.0288$ & $-0.0367$ & $-0.0244$ & $-0.0365$ & $-0.0247$ & $-0.0298$ & $-0.0272$ \\
\bottomrule
\end{tabular}
\vspace{-2mm}
\end{table}

\paragraph{\method improves over vanilla MoE at every scale.}
Table~\ref{tab:main_results} shows that the default \methodlock reduces vanilla test loss at all five scales, by 0.0244--0.0367 at 30B tokens and by 0.0247 at 1.5B after 60B tokens, and by 0.0298 and 0.0272 under the 16E/top-2 and 32E/top-4 budgets.
Relative to persistent full-pool training, the lock changes loss by $-0.0021$ to $+0.0064$ across the six settings where both exist.
\methodfull itself improves over vanilla at all four 30B-token scales, by 0.0288--0.0386.
At the four 8E/top-1 scales with matched dense baselines, all MoE variants reach far lower test loss than the dense models.

\paragraph{The gap holds throughout training.}
\methodfull stays below vanilla MoE after warmup at 182M, 469M, and 650M, and the gap opens early rather than only at the final checkpoint (Appendix Figure~\ref{fig:training_curves}).
The sharing-scope runs follow the same ordering over training as at the end, with the global pool lowest and the layer-private model highest (panel~\subref{fig:pool_size}).

\begin{table}[t]
\centering
\caption{Zero-shot downstream accuracy (\%, raw \texttt{acc}) of vanilla MoE, \methodfull, and \methodlock. The bottom block adds full-width experts to the 12-layer 182M backbone, giving about 268M and 438M active parameters. \methodfull was not trained at 1.5B. \textbf{Bold} marks the best method for each scale and task; Avg is the unweighted mean.}
\label{tab:downstream}
\vspace{-2pt}
\resizebox{\textwidth}{!}{%
\begin{tabular}{@{}l l l ccccccc c@{}}
\toprule
\textbf{Setting} & \textbf{Scale} & \textbf{Method} & \textbf{ARC-E} $\uparrow$ & \textbf{ARC-C} $\uparrow$ & \textbf{PIQA} $\uparrow$ & \textbf{HellaSwag} $\uparrow$ & \textbf{WinoGrande} $\uparrow$ & \textbf{LAMBADA} $\uparrow$ & \textbf{RACE} $\uparrow$ & \textbf{Avg} $\uparrow$ \\
\midrule
\multirow{3}{*}{8E / top-1} & \multirow{3}{*}{182M}
 & Vanilla MoE & 45.71 & 19.97 & 63.11 & 29.98 & \textbf{50.99} & 32.78 & 28.61 & 38.74 \\
 & & \methodfull    & \textbf{46.72} & 20.48 & \textbf{64.36} & 30.66 & \textbf{50.99} & \textbf{34.56} & \textbf{29.47} & \textbf{39.61} \\
 & & \methodlock    & 46.58 & \textbf{20.56} & 64.14 & \textbf{30.77} & 50.81 & 34.43 & 29.35 & 39.52 \\
\cmidrule(l){2-11}
\multirow{3}{*}{8E / top-1} & \multirow{3}{*}{469M}
 & Vanilla MoE & 50.51 & 21.08 & 66.32 & 32.72 & 51.14 & 40.21 & 29.38 & 41.62 \\
 & & \methodfull    & \textbf{53.16} & 21.42 & \textbf{67.30} & \textbf{33.90} & 52.72 & \textbf{42.19} & \textbf{31.10} & \textbf{43.11} \\
 & & \methodlock    & 52.88 & \textbf{21.57} & 66.94 & 33.71 & \textbf{52.85} & 41.82 & 30.88 & 42.95 \\
\cmidrule(l){2-11}
\multirow{3}{*}{8E / top-1} & \multirow{3}{*}{650M}
 & Vanilla MoE & 51.94 & 21.25 & 67.03 & 34.53 & \textbf{53.04} & 43.74 & 29.76 & 43.04 \\
 & & \methodfull    & 52.02 & \textbf{22.61} & 67.90 & 35.55 & 52.49 & 44.28 & 31.67 & 43.79 \\
 & & \methodlock    & \textbf{52.17} & 22.52 & \textbf{68.12} & \textbf{35.68} & 52.41 & \textbf{44.46} & \textbf{31.80} & \textbf{43.88} \\
\cmidrule(l){2-11}
\multirow{3}{*}{8E / top-1} & \multirow{3}{*}{830M}
 & Vanilla MoE & 52.53 & 23.89 & \textbf{68.93} & 35.36 & 52.33 & 43.14 & 30.53 & 43.82 \\
 & & \methodfull    & \textbf{56.57} & 25.00 & 68.77 & \textbf{36.90} & 52.49 & \textbf{47.37} & \textbf{32.63} & \textbf{45.67} \\
 & & \methodlock    & 56.12 & \textbf{25.18} & 68.22 & 36.58 & \textbf{52.64} & 46.74 & 32.25 & 45.39 \\
\cmidrule(l){2-11}
\multirow{2}{*}{8E / top-1} & \multirow{2}{*}{1.5B}
 & Vanilla MoE & 56.99 & 25.34 & 71.16 & 39.30 & 55.17 & 50.98 & 33.59 & 47.50 \\
 & & \methodlock    & \textbf{58.50} & \textbf{25.85} & \textbf{72.31} & \textbf{40.49} & \textbf{56.27} & \textbf{52.07} & \textbf{33.63} & \textbf{48.45} \\
\midrule
\multirow{3}{*}{16E / top-2} & \multirow{3}{*}{268M}
 & Vanilla MoE & 48.82 & \textbf{21.59} & 65.07 & 31.83 & 49.72 & 36.48 & 28.80 & 40.33 \\
 & & \methodfull    & \textbf{49.24} & 20.22 & \textbf{65.45} & \textbf{32.33} & 54.22 & \textbf{37.86} & \textbf{29.19} & \textbf{41.22} \\
 & & \methodlock    & 48.93 & 20.34 & 65.03 & 32.14 & \textbf{54.37} & 37.38 & 28.88 & 41.01 \\
\cmidrule(l){2-11}
\multirow{3}{*}{32E / top-4} & \multirow{3}{*}{438M}
 & Vanilla MoE & 50.08 & 21.08 & 66.43 & 32.91 & \textbf{51.54} & 39.41 & 29.00 & 41.49 \\
 & & \methodfull    & 52.44 & \textbf{22.27} & 67.41 & 34.32 & 50.51 & 40.77 & 30.62 & 42.62 \\
 & & \methodlock    & \textbf{52.60} & 22.19 & \textbf{67.62} & \textbf{34.44} & 50.45 & \textbf{40.92} & \textbf{30.75} & \textbf{42.71} \\
\bottomrule
\end{tabular}%
}
\vspace{-3mm}
\end{table}

\paragraph{Downstream accuracy follows loss.}
\label{sec:downstream}
We evaluate ARC-Easy and ARC-Challenge~\citep{clark2018think}, PIQA~\citep{bisk2020piqa}, HellaSwag~\citep{zellers2019hellaswag}, WinoGrande~\citep{sakaguchi2021winogrande}, LAMBADA~\citep{paperno2016lambada}, and RACE~\citep{lai2017race} zero-shot (Table~\ref{tab:downstream}).
\methodfull raises average accuracy by 0.87, 1.49, 0.75, and 1.85 points from 182M to 830M, winning 25 of 28 scale--task cells; ARC-E, HellaSwag, and LAMBADA improve at every scale, by up to 4.23 points on LAMBADA at 830M.
\methodlock tracks \methodfull downstream as it does in loss: its average stays within 0.3 points of \methodfull in the six settings where both were evaluated, and it raises average accuracy over vanilla MoE by 0.68--1.57 points across seven settings, winning 43 of 49 cells.
At 1.5B, \methodlock raises average accuracy from 47.50 to 48.45 and improves all seven tasks.

\paragraph{Post-lock throughput matches vanilla MoE.}
\label{sec:training_throughput}
Table~\ref{tab:throughput} separates the two phases of each \methodlock run.
Dispatching over the full pool costs 22\%--36\% more per step than vanilla MoE.
After the lock, compact dispatch cuts step time by 16\%--27\% and brings it to within $-0.7\%$ to $+2.8\%$ of vanilla, so the end-to-end overhead of \methodlock, weighting the two phases by their 2K and 58K steps, is $+0.6\%$ to $+3.5\%$ while its loss stays close to \methodfull (Table~\ref{tab:main_results}).
The residual post-lock overhead is the cost of scoring the global $M$-expert index and pool statistics, which we retain so that the objective is unchanged after the lock.

\subsection{Where Does the Gain Come From?}
\label{sec:ownership}
\label{sec:early_full_pool}

\begin{figure}[t]
\centering
\begin{minipage}[t]{0.40\textwidth}
\vspace{0pt}
\centering
\captionsetup{justification=raggedright,singlelinecheck=false}
\captionof{table}{182M test loss (8E/top-1) by when ownership is decided. A and B share one random allocation; C and \methodlock are two seeds of the same recipe. All controls: Appendix Table~\ref{tab:ownership_full}.}
\label{tab:ownership}
\small
\setlength{\tabcolsep}{4pt}
\begin{tabular}{@{}lc@{}}
\toprule
\textbf{Run} & \textbf{Loss} $\downarrow$ \\
\midrule
\multicolumn{2}{@{}l}{\textit{Fixed at initialization}} \\
Vanilla MoE & 1.9317 \\
A: random disjoint & 1.9319 \\
Routed shared core & 1.9262 \\
ReXMoE-R2 (adjacent reuse) & 1.9325 \\
MoUE-stateless (windowed) & 1.9411 \\
\midrule
\multicolumn{2}{@{}l}{\textit{Global access, then locked}} \\
B: random allocation & 1.9173 \\
C: learned allocation & 1.9040 \\
\methodlock (first seed) & \textbf{1.9017} \\
\midrule
\multicolumn{2}{@{}l}{\textit{Never fixed}} \\
\methodfull & 1.9029 \\
\bottomrule
\end{tabular}
\end{minipage}\hfill
\begin{minipage}[t]{0.56\textwidth}
\vspace{0pt}
\centering
\includegraphics[width=\linewidth]{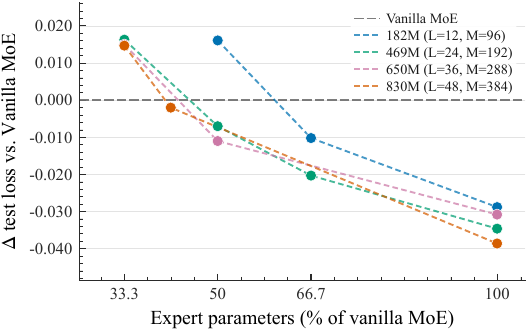}
\vspace{-5mm}
\captionsetup{justification=raggedright,singlelinecheck=false}
\captionof{figure}{Reduced pools: change in final test loss of \methodfull relative to vanilla MoE with eight experts per layer (vertical axis) versus pool size as a fraction of vanilla's expert parameters, at unchanged top-1 active compute.}
\label{fig:efficiency_sweeps}
\end{minipage}
\vspace{-2mm}
\end{figure}

\paragraph{An allocation fixed before training carries no information.}
Table~\ref{tab:ownership} groups 182M runs by when the layer-to-expert allocation is decided.
At initialization the experts are exchangeable, so a disjoint allocation fixed at step 0 is vanilla MoE up to a relabeling of experts, whichever experts it gives to which layer.
Control A confirms this and serves as a sanity check for the other controls: it uses \method's pool, router, and losses, and differs from \methodlock only in drawing its disjoint allocation at random at step 0, yet it matches vanilla to within 0.0002.
Control A therefore isolates the gain to full-pool training itself rather than to the pool, the router, or the losses on their own.
Sharing patterns fixed before training, including routed shared cores, adjacent-layer reuse, and windowed universal experts, are not equivalent to vanilla, yet they also land between 1.9262 and 1.9411, around vanilla's 1.9317.
The strongest fixed pattern, a routed shared core, improves on vanilla by 0.0055 at 182M and 0.0167 at 469M, less than half of \method's gain; a DeepSeekMoE-style shared expert under the 16E/top-2 budget behaves similarly (Appendix Table~\ref{tab:expert_reuse_baselines}).

\paragraph{Locking the learned allocation recovers the full-pool gain.}
Control C trains with the full pool, fixes the learned allocation at 1K, attenuates the other experts, and locks it at 2K, exactly as \methodlock does.
It improves on A by 0.0279 and matches \methodfull (1.9029): brief full-pool training followed by the lock recovers the gain of persistent full-pool training.
Control B isolates the allocation itself by following the same schedule but locking A's random allocation instead; it improves on A by 0.0146, so discarding the learned allocation at the lock forfeits about half of the gain, and retaining the learned allocation provides the remaining improvement.
Since B and C differ only in the locked allocation, their gap isolates the learned allocation's value given the shared phase, not the phase's effect on the expert weights themselves, which a matched warm-start control would test.
C and the reported \methodlock run are two seeds of the same recipe (1.9040 and 1.9017), whose 0.0023 spread matches the seed standard deviation at 182M.
Consistent with the small gap between \methodlock and \methodfull, persistent cross-layer reuse in \methodfull is limited: a token touches 94.1\% distinct experts across 12 layers, 89.5\% across 24, and 83.4\% across 36, against 94.5\%, 94.2\%, and 94.2\% under independent uniform routing, so reuse beyond chance appears only in the deeper models (Appendix~\ref{app:distinct_expert_accounting}).

\begin{table}[t]
\centering
\begin{minipage}[t]{0.53\textwidth}
\vspace{0pt}
\centering
\captionsetup{justification=raggedright,singlelinecheck=false,position=top,skip=4pt}
\captionof{table}{Ablations of \methodfull at 182M. $\Delta$ is relative to vanilla MoE; $G$: number of expert pools.}
\label{tab:ablation}
\footnotesize
\renewcommand{\arraystretch}{0.92}
\setlength{\tabcolsep}{5pt}
\begin{tabular}{@{}lcc@{}}
\toprule
\textbf{Configuration} & \textbf{Loss} $\downarrow$ & $\Delta$ \\
\midrule
\multicolumn{3}{@{}l}{\textit{Components}} \\
Vanilla MoE + softmax ($G{=}12$) & 1.9317 & -- \\
Vanilla MoE + NormRouter & 1.9375 & $+$0.0058 \\
Vanilla MoE, sigmoid, aux-free & 1.9239 & $-$0.0078 \\
Shared + layer aux + softmax & 1.9480 & $+$0.0163 \\
Shared + pool aux + softmax & 1.9180 & $-$0.0137 \\
\methodfull ($G{=}1$) & \textbf{1.9029} & $-$0.0288 \\
\midrule
\multicolumn{3}{@{}l}{\textit{Sharing scope}} \\
$G{=}6$ & 1.9121 & $-$0.0196 \\
$G{=}4$ & 1.9099 & $-$0.0218 \\
$G{=}2$ & 1.9213 & $-$0.0104 \\
\bottomrule
\end{tabular}
\end{minipage}\hfill
\begin{minipage}[t]{0.43\textwidth}
\vspace{0pt}
\centering
\captionsetup{justification=raggedright,singlelinecheck=false,position=top,skip=4pt}
\captionof{table}{Step-time overhead of \methodlock relative to vanilla MoE before the lock (\emph{Full}), after it (\emph{Compact}), and end-to-end over 60K steps (\emph{E2E}), on 8 GPUs. Raw step times: Table~\ref{tab:throughput_full}.}
\label{tab:throughput}
\label{tab:curriculum_throughput}
\footnotesize
\setlength{\tabcolsep}{5pt}
\begin{tabular}{@{}lrrr@{}}
\toprule
\textbf{Setting} & \textbf{Full} & \textbf{Compact} & \textbf{E2E} \\
\midrule
182M, 8E  & $+22\%$ & $+2.8\%$ & $+3.5\%$ \\
182M, 16E & $+25\%$ & $+1.3\%$ & $+2.1\%$ \\
182M, 32E & $+36\%$ & $-0.7\%$ & $+0.6\%$ \\
469M, 8E  & $+33\%$ & $+2.5\%$ & $+3.5\%$ \\
\bottomrule
\end{tabular}
\end{minipage}
\vspace{-4mm}
\end{table}

\subsection{What the Full Pool Reveals}
\label{sec:full_pool_analysis}

\paragraph{Reduced pools improve parameter efficiency.}
\label{sec:param_efficiency}
Figure~\ref{fig:efficiency_sweeps} is a \methodfull parameter-efficiency study: it shrinks the shared pool below $8L$ while keeping one active expert per layer and token.
Pools with 66.7\% of vanilla's expert parameters outperform vanilla at 182M (1.9215 vs.\ 1.9317), and pools with 50\% outperform it at 469M and 650M by 0.007 and 0.011.
At 182M, we directly compare the 50\% shared pool ($M{=}48$) with a parameter-matched private baseline using four private experts per layer: the shared pool reaches 1.9479 versus 1.9664 for the private baseline under the same global batch size, data, schedule, and seed.
At 830M a pool with 41.7\% of the expert parameters matches vanilla to within 0.002.
Across these jointly scaled configurations, the smallest competitive fraction decreases with depth.

\paragraph{More experts and looped execution.}
\methodfull improves over vanilla MoE at 8E/top-1, 16E/top-2, and 32E/top-4 on the 182M backbone (Table~\ref{tab:main_results}), so the gain persists when each layer stores and activates more full-width experts.
With two recurrent expert steps per layer, \methodfull reduces loss from 1.9077 to 1.8727 at 182M and from 1.7768 to 1.7389 at 469M (Appendix~\ref{app:looped_expert}).
In this looped variant, attention runs once per layer and the routed expert sublayer is applied twice with the same router and experts, so expert parameters and per-token expert compute stay matched.

\begin{figure}[t]
\centering
\begin{subfigure}[t]{0.46\textwidth}
    \centering
    \includegraphics[width=\textwidth]{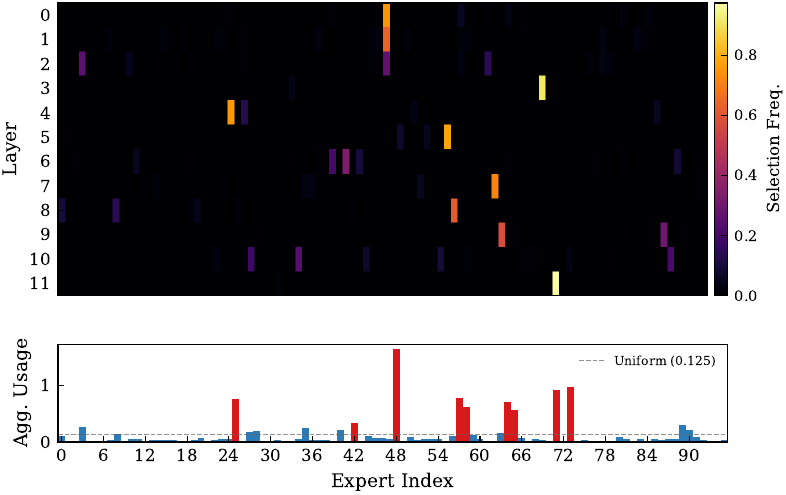}
    \caption{Softmax + per-layer aux loss}
\end{subfigure}\hfill
\begin{subfigure}[t]{0.46\textwidth}
    \centering
    \includegraphics[width=\textwidth]{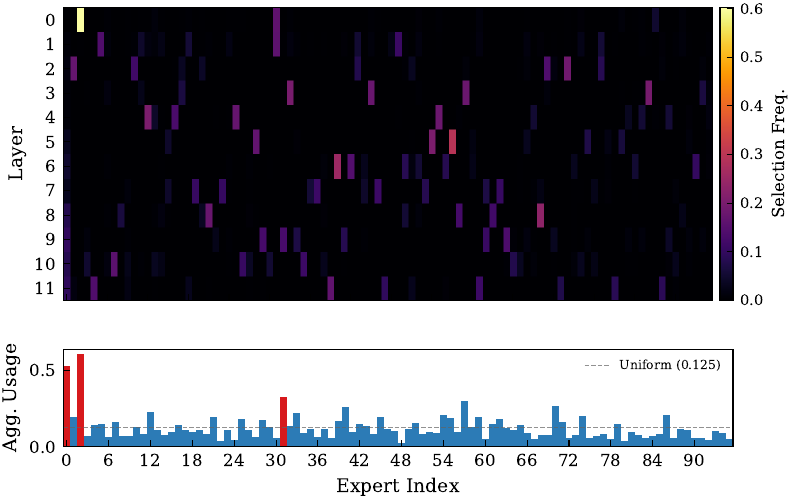}
    \caption{\methodfull: NormRouter + pool aux loss}
\end{subfigure}
\caption{Expert usage in a 182M shared pool, per layer (top) and aggregated (bottom). Bottom panels show aggregate usage summed across layers (uniform $L/M=1/E=0.125$; red bars exceed it). (a) Per-layer balancing collapses aggregate usage onto a few experts. (b) The pool loss with NormRouter balances aggregate usage while each layer concentrates on its own subset of experts.}
\label{fig:load_balance}
\vspace{-3mm}
\end{figure}

\paragraph{Component ablations.}
Table~\ref{tab:ablation} decomposes \methodfull at 182M.
Sharing the pool with per-layer balancing and softmax routing is worse than vanilla (1.9480 vs.\ 1.9317) and collapses aggregate usage onto a few experts (Figure~\ref{fig:load_balance}a).
Replacing it with the pool loss improves loss to 1.9180, and NormRouter improves it further to 1.9029.
With both, aggregate usage is balanced while each layer concentrates on its own subset of the pool (Figure~\ref{fig:load_balance}a--b), the layer--expert preference that the lock later turns into ownership.
NormRouter does not help private experts (1.9375), so its benefit depends on the shared candidate set.
\methodfull also beats the stronger private sigmoid-gating, auxiliary-loss-free baseline at 182M (1.9239) by 0.0210.
Grouping layers into $G$ pools interpolates between private ($G{=}12$) and global ($G{=}1$) sharing: every intermediate $G$ beats vanilla and the global pool is best.

\begin{wraptable}{r}{0.46\textwidth}
\vspace{-12pt}
\centering
\small
\setlength{\tabcolsep}{4pt}
\caption{Single-layer routing randomization on our 469M models: deep-half average over five tasks, length-normalized accuracy (so \textsc{Top-K} differs from Table~\ref{tab:downstream}). $^\dagger$: sampled from each layer's eight most-used experts.}
\label{tab:own_model_randomization}
\begin{tabular}{@{}lccc@{}}
\toprule
\textbf{Model} & \textsc{Top-K} & \textsc{Random} & $\Delta$ \\
\midrule
Vanilla MoE & 45.10 & 43.83 & $-1.3$ \\
\methodlock & 47.05 & 43.77 & $-3.3$ \\
\methodfull & 47.16 & 43.10$^\dagger$ & $-4.1$ \\
\bottomrule
\end{tabular}
\vspace{-6pt}
\end{wraptable}
\paragraph{Learned ownership makes experts less substitutable.}
\label{sec:routing_sensitivity}
\label{sec:homogeneity_own_models}
Repeating the probe of Section~\ref{sec:observation} on our models (Table~\ref{tab:own_model_randomization}), randomizing one deep-half layer costs vanilla MoE 1.3 points at 469M, matching the production models.
We match the randomized choice-set sizes across models; the full protocol is given in Appendix~\ref{app:analysis_details}.
The same probe costs \methodlock 3.3 points, although each of its layers holds exactly as many experts as a vanilla layer and the random choice is drawn from that locked allocation.
It costs \methodfull 4.1 points when sampling from each layer's eight most-used experts, and a similar 3.8 when sampling from the entire pool.
With matched choice-set sizes, the larger drop is consistent with stronger specialization under global competition; because gates are re-normalized at intervention, the probe is a sensitivity test rather than a causal test of ownership.

\section{Conclusion}
\label{sec:conclusion}
\methodlock learns which experts each layer owns through brief global access, then locks this allocation and trains a layer-private MoE.
Across five scales from 182M to 1.5B, the learned allocation improves over vanilla MoE and recovers nearly all of the gain of persistent full-pool training, with post-lock step time within $+2.8\%$ of vanilla (Table~\ref{tab:throughput}).
The persistent full-pool reference also beats vanilla at 50\%--66.7\% of its expert-parameter budget.
Appendix~\ref{sec:limitations} discusses limitations.

\newpage
\subsection*{AI use statement}
Generative AI tools were used to assist with manuscript language editing and LaTeX format conversion.
They were not used to generate experimental results or alter empirical data.
The authors reviewed the AI-assisted text and formatting, verified the manuscript against the underlying experiments and source material, and take responsibility for the final content of this work.

\subsection*{Reproducibility statement}
Section~\ref{sec:setup} describes the experimental setup, and Appendices~\ref{app:model_configs} and~\ref{app:hyperparams} report model configurations and training hyperparameters.
Appendix~\ref{app:progressive_curriculum_details} specifies the ownership-learning schedule, assignment, and compact-dispatch invariants, and Appendix~\ref{app:curriculum_mechanism} specifies the ownership controls.
Appendices~\ref{app:expert_reuse_baselines}, \ref{app:looped_expert}, \ref{app:analysis_details}, \ref{app:pool_aux_derivation}, and \ref{app:monte_carlo} cover the sharing baselines, looped experts, evaluation protocols, pool-loss implementation, and NormRouter initialization.

\bibliography{references}
\bibliographystyle{preprint}

\newpage
\appendix

\section{Alternative Expert-Sharing Baselines}
\label{app:expert_reuse_baselines}

We compare alternative sharing mechanisms under the two expert budgets used in the main experiments: 8E/top-1 for the 182M and 469M backbones, and 16E/top-2 on the 182M backbone.
All controls use the corresponding LLaMA backbone, Pile token budget, 60K-step schedule, and global batch size 512.
The added controls are single runs with seed 1234 and micro-batch size 64.
We match the number and width of stored expert FFNs and the number of activated expert FFNs per token.
We treat the small router deltas (at most 0.18\% here) as the same total-parameter budget while reporting their exact size below.
Table~\ref{tab:ownership_full} lists the full 182M ownership comparison summarized in Table~\ref{tab:ownership}.

\begin{table}[h]
\centering
\caption{Fixed versus learned expert ownership at 182M under the 8E/top-1 expert budget (96 expert FFNs, one active per layer and token). Test loss after 60K steps; lower is better. C and \methodlock are two seeds of the same recipe. Rows A--C use the same NormRouter, pool loss, and cardinality regularizer and end with exactly eight disjoint experts per layer; A and B share the same random allocation. Vanilla MoE and \methodfull average three seeds; all other rows are single runs (protocol in Appendix~\ref{app:expert_reuse_baselines}).}
\label{tab:ownership_full}
\vspace{4pt}
\small
\begin{tabular}{@{}llc@{}}
\toprule
\textbf{Ownership} & \textbf{Configuration} & \textbf{Loss} $\downarrow$ \\
\midrule
\multirow{5}{*}{Fixed at initialization}
 & Vanilla MoE (private banks) & 1.9317 \\
 & A: random disjoint allocation, same losses as \method & 1.9319 \\
 & Routed shared core (4 shared + private) & 1.9262 \\
 & ReXMoE-R2~\citep{tan2025rexmoe} (adjacent-layer reuse) & 1.9325 \\
 & MoUE-stateless~\citep{chen2026moue} (private + windowed universal) & 1.9411 \\
\midrule
Global access, then random & B: full pool, then A's allocation locked at 2K & 1.9173 \\
\midrule
\multirow{2}{*}{Learned, then locked}
 & C: full pool, then learned allocation locked at 2K & 1.9040 \\
 & \methodlock (first seed) & \textbf{1.9017} \\
\midrule
Never fixed & \methodfull & 1.9029 \\
\bottomrule
\end{tabular}
\end{table}

\paragraph{Routed shared core.}
We use \emph{Routed Shared Core} as a descriptive name for this control.
It installs a seed-1234 static random allocation before the first optimization step, with four core experts visible at every layer and fixed specialist assignments for the remainder of training.
The core experts remain routing candidates rather than always-active branches.
Assigning every remaining specialist to exactly one layer yields candidate-set sizes of 11 or 12: eight 12-expert and four 11-expert layers at 182M, and twenty 12-expert and four 11-expert layers at 469M.
This topology is closest to MoUE's mixture of universal and private experts~\citep{chen2026moue}, but here one four-expert routed core spans every depth, the other experts are layer-private, and the allocation and routers are stateless.
Unlike \methodlock, it retains four shared candidates throughout training and never learns an allocation from full-pool access.
The 182M and 469M variants contain global pools of 96 and 192 physical expert FFNs, respectively, and use top-1 routing, exactly matching vanilla MoE's expert-FFN storage and per-token expert activation.
Their independent layer routers retain the full 96- or 192-way global coordinates, so total stored parameters include the same router overhead as \methodfull: 0.81M at 182M and 4.52M at 469M (0.10\% and 0.18\% of the corresponding vanilla models).
Compact routing restricts the post-logit normalization and top-$k$ operations to the fixed candidate set, but it does not remove rows from the stored gate or its full-pool linear projection.

\paragraph{ReXMoE-R2.}
We group two adjacent layers and expose each router to the union of their experts, retaining top-1 softmax routing and per-layer balancing.
Each router therefore accesses 16 candidates, progressively expanded from 8 to 16 between steps 10K and 30K.
The wider candidate set adds a small router-parameter and compute overhead beyond the matched expert FFN budget.

\paragraph{MoUE-stateless.}
Our parameter-matched MoUE adaptation has 96 physical expert FFNs: four private experts per layer (48 total) and 48 universal experts.
Each layer routes top-1 over its four private experts and a stride-4 rotating window of eight universal experts, so every universal expert is visible at two depths.
Following MoUE, we use exposure-corrected local/universal balancing ($\alpha_{\mathrm{local}}{=}1$, $\alpha_{\mathrm{univ}}{=}0.5$) and add a universal-expert logit bias of 0.75 that decays linearly over the first 5\% of training.
We retain independent per-layer softmax routers and omit the trajectory-state Universal Router, and label the result \emph{MoUE-stateless}.
Like \methodfull, this adaptation uses wider global-coordinate routers than vanilla MoE, but all three methods store 96 expert FFNs and activate one expert per layer and token.

\paragraph{DeepSeekMoE-style shared expert.}
For the 182M 16E/top-2 budget, this control replaces vanilla MoE's 16 routed experts with 15 routed experts selected by top-1 softmax routing and one always-active shared expert at each layer, following the shared-expert mechanism of DeepSeekMoE~\citep{dai2024deepseekmoe}.
The shared expert is layer-local rather than shared across depth.
Both configurations therefore store 16 expert FFNs per layer and activate two expert FFNs per token and layer.
The shared-expert control's router has 15 rather than 16 output logits, reducing the total stored parameter count by only $12\times768=9{,}216$ router weights; expert parameters and expert activation are exactly matched.
We use a per-layer auxiliary-loss coefficient of 0.01, as in the matched vanilla 16E/top-2 run.
The shared-expert run uses micro-batch size 64 and the matched vanilla 16E/top-2 run uses 16; both use seed 1234, global batch size 512, and the same 60K-step schedule.
We treat this as a matched expert-budget control.

\begin{table}[H]
\centering
\small
\setlength{\tabcolsep}{12pt}
\caption{Test loss for alternative expert-sharing controls. Vanilla MoE and \methodfull values repeat the corresponding rows in the main experiments; the compact method labels are defined above. Added controls are single-seed runs; a dash denotes a setting that was not run.}
\label{tab:expert_reuse_baselines}
\begin{tabular}{lcc}
\toprule
Method & 182M & 469M \\
\midrule
\multicolumn{3}{l}{\emph{8E/top-1 expert budget}} \\
Vanilla MoE & 1.9317 & 1.7982 \\
Routed core (4) & 1.9262 & 1.7815 \\
ReXMoE-R2~\citep{tan2025rexmoe} & 1.9325 & -- \\
MoUE-stateless~\citep{chen2026moue} & 1.9411 & -- \\
\methodfull & \textbf{1.9029} & \textbf{1.7636} \\
\midrule
\multicolumn{3}{l}{\emph{16E/top-2 expert budget}} \\
Vanilla MoE & 1.8589 & -- \\
DS-MoE shared (1)~\citep{dai2024deepseekmoe} & 1.8509 & -- \\
\methodfull & \textbf{1.8277} & -- \\
\bottomrule
\end{tabular}
\end{table}

\section{Limitations and Future Work}
\label{sec:limitations}

\paragraph{Lock schedule.}
All \methodlock runs use one schedule: allocation at 1K, cosine attenuation over 1K--2K, and lock at 2K, with cardinality target $K_{\mathrm{tar}}=E$.
We do not ablate the allocation or lock times, the attenuation schedule, or the cardinality coefficient; sensitivity to this schedule and its tuning per scale are left to future work.

\paragraph{Full-pool results.}
Reduced pools and looped execution are reported for \methodfull.
Combining reduced pools with the lock requires smaller or non-uniform per-layer allocations and is left to future work.
Our alternative-sharing controls are parameter-matched adaptations (Appendix~\ref{app:expert_reuse_baselines}).

\paragraph{Downstream evaluation.}
Evaluation is zero-shot on seven benchmarks; few-shot and post-training evaluation are future work.

\clearpage
\section{Model and MoE Configurations}
\label{app:model_configs}

\begin{table}[h]
\centering
\caption{Backbone configurations for the four 30B-token scales. All models use sequence length 1024, dense-width FFNs (or expert FFNs for MoE variants) with intermediate size ($4 \times H$), and SwiGLU activation. Vanilla MoE and \method are matched in total expert FFNs and per-token expert FLOPs; stored and active parameter counts for both are listed in Table~\ref{tab:full_hyperparams}, and the small deltas come from the larger shared-pool routers in \method, which are identical for \methodfull and \methodlock. The 16E/top-2 and 32E/top-4 variants of the 182M backbone keep full-width experts (7.08M parameters each) and have 1{,}458M and 2{,}819M total and about 269M and 441M active parameters for \method. The 1.5B backbone is described in Appendix~\ref{app:hyperparams}.}
\label{tab:model_configs}
\vspace{4pt}
\small
\begin{tabular}{lccccccc}
\toprule
\textbf{Scale} & \textbf{L} & \textbf{H} & \textbf{FFN} & \textbf{Heads} & \textbf{KV} & \textbf{$E$/layer} & \textbf{$E_{\mathrm{pool}}$} \\
\midrule
182M & 12 & 768 & 3072 & 12 & 4 & 8 & 96 \\
469M & 24 & 1024 & 4096 & 16 & 4 & 8 & 192 \\
650M & 36 & 1024 & 4096 & 16 & 4 & 8 & 288 \\
830M & 48 & 1024 & 4096 & 16 & 4 & 8 & 384 \\
\bottomrule
\end{tabular}
\end{table}

\begin{table}[h]
\centering
\caption{MoE configuration comparison between vanilla MoE and \method. The two are matched in total expert FFNs and per-token expert FLOPs. The ownership lock of \methodlock is specified in Appendix~\ref{app:progressive_curriculum_details}.}
\label{tab:moe_configs}
\vspace{4pt}
\begin{tabular}{lcc}
\toprule
 & \textbf{Vanilla MoE} & \textbf{\method} \\
\midrule
Expert ownership & Per-layer & Global shared pool \\
Number of experts & 8 per layer & $8 \times L$ (global pool) \\
Total expert FFNs & $8L$ & $8L$ \\
Expert evals per token & $L$ & $L$ \\
Routing & Top-1, softmax & Top-1, NormRouter \\
Per-layer aux loss & $1 \times 10^{-2}$ & 0 \\
Pool aux loss & --- & $1$--$2 \times 10^{-2}$ \\
Grouped GEMM & \checkmark & \checkmark \\
\bottomrule
\end{tabular}
\end{table}

\clearpage
\section{Looped-Expert Compatibility Details}
\label{app:looped_expert}

This appendix expands the looped-expert result summarized in Section~\ref{sec:main_results}.
Motivated by recurrent Transformer designs~\citep{dehghani2019universal,csordas2024moeut}, the experiment tests whether \methodfull remains effective when the routed expert computation is reused recurrently within each layer.

\paragraph{Recurrent expert computation.}
Let $a_l$ be the residual-stream state after the attention sublayer of layer $l$, and set $z_l^{(0)}=a_l$.
For $R=2$ expert steps, we compute
\begin{equation}
    u_l^{(r)} = \operatorname{LN}_{l,\mathrm{MLP}}\!\left(z_l^{(r)}\right),
    \qquad
    z_l^{(r+1)} = z_l^{(r)} + \frac{1}{R}\operatorname{MoE}_l\!\left(u_l^{(r)}\right),
    \quad r\in\{0,1\}.
    \label{eq:looped_expert}
\end{equation}
Attention is evaluated once per layer, outside the recurrence in Eq.~\eqref{eq:looped_expert}; only the pre-MLP normalization, router, and expert FFN path is applied twice.
Both steps reuse the same normalization, router, and expert parameters.
The second step nevertheless recomputes its top-1 routing decision from the updated state $z_l^{(1)}$, so a token may select a different expert.
We use neither iteration-specific router biases nor frozen first-step expert assignments.
The factor $1/R$ keeps the aggregate residual-update scale comparable to a single expert step.

\paragraph{Comparison protocol and expert-budget matching.}
The vanilla looped MoE assigns eight private experts to every layer, giving $8L$ physical expert FFNs in total.
The looped \methodfull variant retains independent per-layer routers but replaces those private sets with one global pool of $M=8L$ expert FFNs: $M=96$ for the 12-layer 182M backbone and $M=192$ for the 24-layer 469M backbone.
Both variants use top-1 routing and $R=2$, and therefore execute $2L$ expert FFNs per token.
The comparison matches the number of expert parameters and active expert computation; as in the main experiments, \methodfull has wider router output layers because every layer scores the global pool.
Vanilla uses softmax routing with the standard per-layer auxiliary loss, whereas \methodfull uses NormRouter and the pool-level auxiliary loss.
All four runs use the corresponding backbone from Table~\ref{tab:model_configs}, a global batch size of 512, sequence length 1,024, and the 60K-step training schedule in Appendix~\ref{app:hyperparams}.

\begin{table}[h]
\centering
\caption{Looped-expert compatibility after 60K training steps. All runs use $R=2$ recurrent expert steps and top-1 routing. ``Experts'' gives the physical expert allocation; $\Delta$ loss is \methodfull minus vanilla MoE within a scale. Lower is better.}
\label{tab:looped_expert}
\vspace{4pt}
\small
\resizebox{\textwidth}{!}{%
\begin{tabular}{@{}llccccccc@{}}
\toprule
\textbf{Scale} & \textbf{Backbone ($L/H$)} & \textbf{Vanilla experts} & \textbf{\methodfull experts}
& \multicolumn{2}{c}{\textbf{Vanilla MoE}} & \multicolumn{2}{c}{\textbf{\methodfull}} & \textbf{$\Delta$ loss} \\
\cmidrule(lr){5-6}\cmidrule(lr){7-8}
& & & & \textbf{Loss} $\downarrow$ & \textbf{PPL} $\downarrow$ & \textbf{Loss} $\downarrow$ & \textbf{PPL} $\downarrow$ & \\
\midrule
182M & 12/768  & 8/layer (96 total)  & 96 global  & 1.9077 & 6.7373 & \textbf{1.8727} & \textbf{6.5061} & $-0.0349$ \\
469M & 24/1024 & 8/layer (192 total) & 192 global & 1.7768 & 5.9106 & \textbf{1.7389} & \textbf{5.6910} & $-0.0379$ \\
\bottomrule
\end{tabular}%
}
\end{table}

Table~\ref{tab:looped_expert} shows that \methodfull improves both loss and perplexity at both tested scales.
The loss reductions of 0.0349 and 0.0379 are comparable to the corresponding single-step reductions of 0.0288 and 0.0346 in the main experiments, indicating that recurrent expert reuse does not erase the benefit of global expert pooling.
This experiment loops the routed expert sublayer rather than the entire Transformer block: attention remains layer-specific and executes once per layer.

\section{Additional Training Curves}
\label{app:training_curves}

\begin{figure}[!htbp]
\centering
\captionsetup{font=small,skip=3pt}
\begin{subfigure}[t]{0.47\textwidth}
    \centering
    \includegraphics[width=\textwidth]{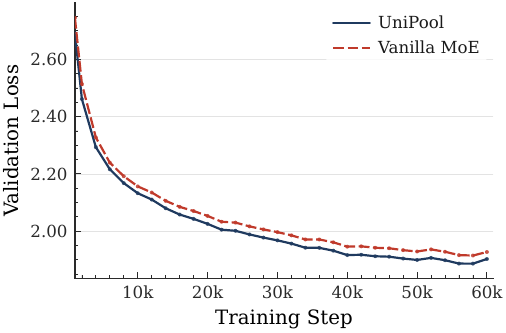}
    \caption{182M}
\end{subfigure}
\hfill
\begin{subfigure}[t]{0.47\textwidth}
    \centering
    \includegraphics[width=\textwidth]{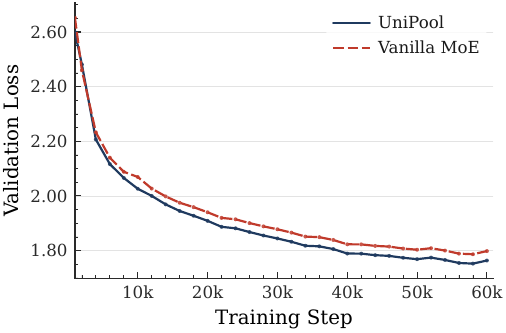}
    \caption{469M}
\end{subfigure}\\[2pt]
\begin{subfigure}[t]{0.47\textwidth}
    \centering
    \includegraphics[width=\textwidth]{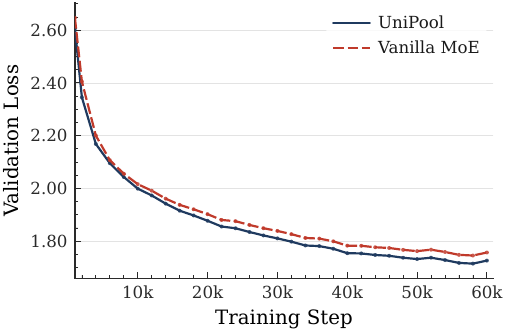}
    \caption{650M}
\end{subfigure}
\hfill
\begin{subfigure}[t]{0.47\textwidth}
    \centering
    \includegraphics[width=\textwidth]{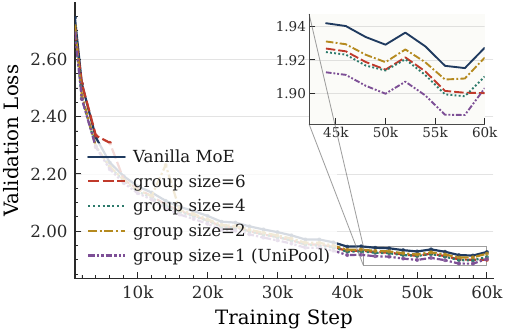}
    \caption{Sharing scope ablation (182M)}
    \label{fig:pool_size}
\end{subfigure}
\vspace{-0.4em}
\caption{\textbf{Validation loss curves.} Panels (a)--(c) compare \methodfull with vanilla MoE at 182M, 469M, and 650M over 30B Pile tokens. Panel (d) shows the 182M sharing-scope ablation, where $G{=}1$ is \methodfull and $G{=}12$ is vanilla MoE; grouped configurations interpolate between the endpoints. Curves are measured on the validation split; the final losses in Table~\ref{tab:main_results} are measured on the held-out test split.}
\label{fig:training_curves}
\end{figure}

Figure~\ref{fig:training_curves} complements the endpoint test losses in Section~\ref{sec:main_results} by showing the full optimization trajectories on the validation split.
Across the 182M, 469M, and 650M scales, \methodfull stays below the matched vanilla MoE baseline after the initial warmup phase, indicating that the gain is present throughout training.
At 182M, the gap opens early and widens steadily; at 469M, the two curves diverge visibly after warmup and end with a validation-loss difference of roughly $0.035$; at 650M, \methodfull continues to maintain a clear advantage throughout training.

Panel~(d) reports the 182M sharing-scope ablation over training.
The trajectory ordering mirrors the endpoint ablation results: global sharing ($G{=}1$) remains the lowest-loss configuration for most of training, vanilla MoE ($G{=}12$) is the highest-loss endpoint, and grouped sharing configurations ($G{=}2,4,6$) generally interpolate between them.
This suggests that broader expert sharing improves the optimization trajectory itself, rather than merely selecting a better final checkpoint.

\clearpage
\section{Hyperparameter Details}
\label{app:hyperparams}

Table~\ref{tab:full_hyperparams} provides complete hyperparameter details for all experimental configurations.

\begin{table}[H]
\centering
\caption{Hyperparameters for the four 30B-token scales. The 1.5B backbone is described below the table.}
\label{tab:full_hyperparams}
\small
\resizebox{\textwidth}{!}{%
\begin{tabular}{lcccc}
\toprule
\textbf{Hyperparameter} & \textbf{182M} & \textbf{469M} & \textbf{650M} & \textbf{830M} \\
\midrule
\multicolumn{5}{l}{\textit{Architecture}} \\
Number of layers & 12 & 24 & 36 & 48 \\
Hidden size & 768 & 1024 & 1024 & 1024 \\
FFN intermediate size & 3072 & 4096 & 4096 & 4096 \\
Attention heads & 12 & 16 & 16 & 16 \\
KV heads (GQA) & 4 & 4 & 4 & 4 \\
Sequence length & 1024 & 1024 & 1024 & 1024 \\
Normalization & RMSNorm & RMSNorm & RMSNorm & RMSNorm \\
Activation & SwiGLU & SwiGLU & SwiGLU & SwiGLU \\
Position embedding & RoPE (base 1M) & RoPE (base 1M) & RoPE (base 1M) & RoPE (base 1M) \\
Stored parameters (M, vanilla MoE) & 777.09 & $2{,}583.93$ & $3{,}823.48$ & $5{,}063.02$ \\
Stored parameters (M, \method) & 777.90 & $2{,}588.46$ & $3{,}833.80$ & $5{,}081.50$ \\
Stored parameter overhead & +0.81M (+0.10\%) & +4.52M (+0.18\%) & +10.32M (+0.27\%) & +18.48M (+0.37\%) \\
Active parameters/token (M, vanilla MoE) & 182.55 & 470.01 & 652.58 & 835.16 \\
Active parameters/token (M, \method) & 183.36 & 474.52 & 662.91 & 853.64 \\
\midrule
\multicolumn{5}{l}{\textit{MoE} (\method)} \\
Experts per layer (vanilla MoE) & 8 & 8 & 8 & 8 \\
Global expert pool size & 96 & 192 & 288 & 384 \\
Total expert FFNs (matched) & 96 & 192 & 288 & 384 \\
Router top-$k$ & 1 & 1 & 1 & 1 \\
Pool aux loss coeff & $1 \times 10^{-2}$ & $2 \times 10^{-2}$ & $2 \times 10^{-2}$ & $2 \times 10^{-2}$ \\
Per-layer aux loss coeff & 0 & 0 & 0 & 0 \\
NormRouter & \checkmark & \checkmark & \checkmark & \checkmark \\
Router init & Monte Carlo & Monte Carlo & Monte Carlo & Monte Carlo \\
\midrule
\multicolumn{5}{l}{\textit{Training}} \\
Global batch size & 512 & 512 & 512 & 512 \\
Micro batch size & 16 & 16 & 16 & 16 \\
Training iterations & 60,000 & 60,000 & 60,000 & 60,000 \\
Total tokens & $\sim$30B & $\sim$30B & $\sim$30B & $\sim$30B \\
Learning rate & $5 \times 10^{-4}$ & $5 \times 10^{-4}$ & $5 \times 10^{-4}$ & $5 \times 10^{-4}$ \\
Min learning rate & $5 \times 10^{-5}$ & $5 \times 10^{-5}$ & $5 \times 10^{-5}$ & $5 \times 10^{-5}$ \\
LR schedule & Cosine & Cosine & Cosine & Cosine \\
Warmup fraction & 0.01 & 0.01 & 0.01 & 0.01 \\
Gradient clipping & 1.0 & 1.0 & 1.0 & 1.0 \\
Precision & bf16 & bf16 & bf16 & bf16 \\
Init std & 0.01 & 0.01 & 0.01 & 0.01 \\
\bottomrule
\end{tabular}
}
\end{table}

All scales use RMSNorm~\citep{zhang2019root}, SwiGLU activation~\citep{shazeer2020glu}, rotary positional embeddings (RoPE)~\citep{su2024roformer}, grouped query attention with 4 KV heads, and untied input/output embeddings.
Training uses Megatron-LM with sequence parallelism and distributed optimizer.
Activation checkpointing with MoE layer recompute is enabled for the 469M, 650M, and 830M scales.
The 1.5B backbone uses 36 layers, hidden size 1,600, FFN size 6,400, 20 query heads, and 4 KV heads, with 8 experts per layer for vanilla MoE and a 288-expert pool for \methodlock, corresponding to about 1.49B active and 9.23B stored vanilla-MoE parameters before the shared-pool router overhead. Both the 1.5B vanilla MoE and \methodlock runs use expert parallelism $\mathrm{EP}=8$; they also use seed 1234, global batch size 1,024, and a 57,220-iteration cosine schedule (about 60B tokens) on the same Pile shard family, and we report their final test losses. \methodlock follows the same allocation, attenuation, and lock schedule as the other scales.

\section{Ownership Lock and Compact Dispatch Details}
\label{app:progressive_curriculum_details}

\paragraph{Schedule.}
In all \methodlock runs, every layer sees the full pool for the first 1K steps.
At step 1K we freeze a disjoint allocation computed from routing statistics, attenuate off-allocation scores with the cosine schedule of Eq.~\eqref{eq:curriculum_attenuation} over steps 1K--2K, and hard-lock the allocation at step 2K.
The pool-level auxiliary loss stays active throughout training, and the target-cardinality regularizer of Eq.~\eqref{eq:target_cardinality} uses $K_{\mathrm{tar}}=E$ ($8$, $16$, and $32$ for the 8E, 16E, and 32E budgets) with $\lambda_{\mathrm{card}}$ warmed to $5\times10^{-3}$ over the first 1K steps.

\paragraph{Assignment.}
The allocation solves a small assignment problem in which each expert has exactly one owner layer and each layer receives at least $E$ experts.
Because the pool contains exactly $M=EL$ experts, these constraints force $K_l=E$ for every layer and assign every expert.
Among feasible allocations, the program maximizes the retained routing mass.

\paragraph{Step times.}
Table~\ref{tab:throughput_full} lists the raw phase-median step times behind Table~\ref{tab:throughput}, measured on 8 GPUs with bfloat16, dropless grouped GEMM, and $\mathrm{EP}=\mathrm{TP}=\mathrm{PP}=1$.
Each row uses one micro-batch size and recomputation setting for all three columns, so overheads are computed within a row.
After the lock, compact dispatch reduces step time by 16\%--27\% relative to the pre-lock phase.

\begin{table}[h]
\centering
\small
\setlength{\tabcolsep}{4pt}
\caption{Phase-median step times (s/it) of \methodlock and vanilla MoE. $M$ is the pool size; end-to-end overhead weights the full-pool phase by 2K and the compact phase by 58K of 60K steps.}
\label{tab:throughput_full}
\begin{tabular}{@{}lrcccrrr@{}}
\toprule
& & \multicolumn{3}{c}{\textbf{s/it} $\downarrow$} & \multicolumn{3}{c}{\textbf{Overhead vs.\ vanilla}} \\
\cmidrule(lr){3-5}\cmidrule(l){6-8}
\textbf{Setting} & $M$ & \textbf{Vanilla} & \textbf{Full pool} & \textbf{Compact} & \textbf{Full pool} & \textbf{Compact} & \textbf{End-to-end} \\
\midrule
182M, 8E/top-1  & 96  & 0.319 & 0.390 & 0.328 & $+22.3\%$ & $+2.8\%$ & $+3.5\%$ \\
182M, 16E/top-2 & 192 & 0.467 & 0.585 & 0.473 & $+25.3\%$ & $+1.3\%$ & $+2.1\%$ \\
182M, 32E/top-4 & 384 & 0.767 & 1.041 & 0.762 & $+35.7\%$ & $-0.7\%$ & $+0.6\%$ \\
469M, 8E/top-1  & 192 & 0.762 & 1.014 & 0.781 & $+33.1\%$ & $+2.5\%$ & $+3.5\%$ \\
\bottomrule
\end{tabular}
\end{table}

\paragraph{Compact-dispatch invariants.}
After the lock, visible global expert indices are sorted before routing probabilities and the dispatch map are gathered into the layer-local coordinate system.
We assert that all off-mask route probabilities and token assignments are zero after the lock.
Each compact expert group references its original expert parameters without copying them.
After the lock each expert is invoked only by its owner layer; the global parameter container does not imply cross-layer execution.
Scores, routing objectives, and diagnostics remain in the global coordinate system; only token permutation, token-count metadata, and grouped GEMM use compact coordinates.

\section{Ownership Controls}
\label{app:curriculum_mechanism}

We examine whether an early full-pool phase provides useful structure for subsequent restricted routing.
The matched training controls use the 182M backbone with $L=12$ layers and $M=96$ experts, top-1 routing, and exactly $K=8$ experts per layer after locking.
This is the allocation that the lock in \methodlock produces at this scale (Appendix~\ref{app:progressive_curriculum_details}); the controls impose it explicitly so that all runs share the same final cardinality.

\subsection{Matched controls for early expert access}
\label{app:curriculum_controls}

Table~\ref{tab:ownership_controls} compares three 60K-step runs with the same architecture, NormRouter, full-$M$ score normalization, pool auxiliary coefficient $10^{-2}$, and target-cardinality coefficient warmed to $5\times10^{-3}$ over 1K steps with target eight.
All use seed 1234, global batch 512, sequence length 1024, learning rate $5\times10^{-4}$, and the same 60K cosine learning-rate schedule with 1\% warmup.
Control A fixes a random balanced disjoint allocation from initialization.
B and C allow full-pool access until 1K, freeze a random or learned allocation, attenuate off-allocation scores over 1--2K, and lock at 2K.
A and B use the same random allocation; C selects its allocation from validation routing counts under the same exclusivity and $K{=}8$ constraints.

\begin{table}[ht]
\centering
\small
\caption{Matched ownership controls at 182M. Test NLL is measured at 60K; all runs use fixed disjoint $K{=}8$ after 2K. Each row is a single training run.}
\label{tab:ownership_controls}
\begin{tabular}{@{}llrr@{}}
\toprule
Control & Allocation schedule & Final NLL $\downarrow$ & $\Delta$ vs.\ A \\
\midrule
A & Random $K{=}8$ from step 0 & 1.9319 & --- \\
B & Full pool $\rightarrow$ random $K{=}8$ at 2K & 1.9173 & $-0.0146$ \\
C & Full pool $\rightarrow$ learned $K{=}8$ at 2K & \textbf{1.9040} & $-0.0279$ \\
\bottomrule
\end{tabular}
\end{table}

All three runs end with the same per-layer cardinality and one owner per expert.
C, which locks the learned allocation, improves on A by 0.0279 and matches \methodfull; B, which locks A's random allocation after the same full-pool phase, improves on A by 0.0146, so discarding the learned allocation at the lock forfeits 0.0133 of the gain.

\section{Distinct-Expert Accounting}
\label{app:distinct_expert_accounting}

In \methodfull, a token visits 94.1\% distinct experts across 12 layers, 89.5\% across 24, and 83.4\% across 36.
At 12 layers this equals the reuse expected from independent uniform routing, whereas at 24 and 36 layers reuse exceeds that baseline.
Persistent reuse is thus modest and grows with depth, consistent with the small gap between \methodlock and \methodfull.

For a token $x$ in an $L$-layer top-1 MoE, let $e_l(x)$ be the expert selected at layer $l$.
In vanilla MoE and in \methodlock after the lock, layers own disjoint experts, so a token always touches exactly $L$ distinct expert tensors.
In \methodfull, layers may select the same expert, and the number of distinct experts touched is
\begin{equation}
    U(x) = \left| \{ e_l(x) : l = 1,\ldots,L \} \right|,
    \qquad 1 \le U(x) \le L.
\end{equation}
Collisions occur even without any deliberate reuse, because $L$ selections are drawn from a finite pool.
If every layer routed independently and uniformly over $M$ experts, the expected count would be $M\left[1-(1-1/M)^L\right]$, which we report as a reference in Table~\ref{tab:unique_expert_accounting}.
Observed values below this reference indicate more reuse than chance.

\begin{table}[h]
\centering
\caption{Distinct experts touched per token under top-1 routing in \methodfull, on the validation set. $M$ is the pool size. ``Uniform'' is the expectation under independent uniform routing.}
\label{tab:unique_expert_accounting}
\vspace{4pt}
\begin{tabular}{lrrrr}
\toprule
Setting & $L$ & $M$ & Observed $U/L$ & Uniform $U/L$ \\
\midrule
Full pool & 12 & 96  & 11.29/12 (94.1\%) & 11.34/12 (94.5\%) \\
Reduced pool & 12 & 64  & 11.46/12 (95.5\%) & 11.02/12 (91.8\%) \\
Reduced pool & 12 & 48  & 11.31/12 (94.3\%) & 10.72/12 (89.3\%) \\
Full pool & 24 & 192 & 21.48/24 (89.5\%) & 22.62/24 (94.2\%) \\
Reduced pool & 24 & 96  & 20.79/24 (86.6\%) & 21.33/24 (88.9\%) \\
Full pool & 36 & 288 & 30.03/36 (83.4\%) & 33.90/36 (94.2\%) \\
Reduced pool & 36 & 128 & 30.12/36 (83.7\%) & 31.49/36 (87.5\%) \\
\bottomrule
\end{tabular}
\end{table}

At 12 layers, reuse in the full pool is close to the uniform reference, and reduced pools touch more distinct experts than chance.
At 24 and 36 layers, tokens reuse experts more often than chance, most clearly for the full 36-layer pool (30.03 distinct experts versus 33.90 expected).
Persistent cross-layer reuse in \methodfull is therefore limited and depth-dependent, which is consistent with the small gap between \methodlock and \methodfull.

\section{Additional Routing-Randomization Details}
\label{app:analysis_details}

\paragraph{Production MoE models: per-task results.}
Table~\ref{tab:observation_production_full} reports per-task downstream accuracy under the single-layer deep-half random-routing intervention for the three production MoE models discussed in Section~\ref{sec:observation}. \textsc{Top-K} denotes the model's original learned top-$k$ router and \textsc{Random} denotes the mean accuracy after randomizing one deep-half MoE layer at a time and averaging across layers; Avg is the unweighted mean and drops are measured relative to \textsc{Top-K}.

\begin{table}[h]
\centering
\caption{Routing redundancy under single-layer randomization in production MoE models. Raw accuracy (\%) is reported on five downstream benchmarks.}
\label{tab:observation_production_full}
\vspace{4pt}
\resizebox{\textwidth}{!}{%
\begin{tabular}{llcccccc}
\toprule
\textbf{Model} & \textbf{Routing} & \textbf{ARC-E} & \textbf{ARC-C} & \textbf{PIQA} & \textbf{HellaSwag} & \textbf{WinoGrande} & \textbf{Avg} \\
\midrule
\multirow{2}{*}{Qwen1.5-MoE}
 & \textsc{Top-K}              & 69.23 & 44.20 & 80.47 & 77.30 & 68.43 & 67.92 \\
 & \textsc{Random}             & 66.76 & 42.19 & 79.07 & 76.08 & 67.34 & 66.29 ($-1.6$) \\
\midrule
\multirow{2}{*}{DeepSeek-V2-Lite}
 & \textsc{Top-K}              & 58.59 & 33.02 & 67.57 & 56.82 & 54.93 & 54.19 \\
 & \textsc{Random}             & 57.23 & 32.08 & 65.88 & 55.41 & 54.57 & 53.03 ($-1.2$) \\
\midrule
\multirow{2}{*}{Qwen3-30B-A3B}
 & \textsc{Top-K}              & 79.50 & 55.97 & 80.79 & 77.70 & 71.11 & 73.02 \\
 & \textsc{Random}             & 78.67 & 54.98 & 79.71 & 76.85 & 70.10 & 72.06 ($-1.0$) \\
\bottomrule
\end{tabular}%
}
\end{table}

\paragraph{Matched randomization for shared experts.}
For vanilla MoE, the random-routing intervention samples uniformly from the 8 private experts owned by the selected layer.
For \methodfull, uniform sampling over the full shared pool would not be comparable, because each layer can choose from $M=L\times 8$ experts rather than from 8 private experts.
We therefore first identify each layer's top-8 most-used shared experts on a held-out Pile validation split, then sample uniformly from that per-layer top-8 set during the intervention.
This keeps the randomized choice set the same size as vanilla MoE while respecting the fact that different \methodfull layers can prefer different regions of the global pool.
We also report the standard full-pool random protocol, where \methodfull samples uniformly from all shared experts.
\methodlock needs no such matching: after the lock each layer routes only within its allocation of 8 experts, and the intervention samples uniformly from that allocation.
After replacing the expert identity, we re-normalize the selected expert's gate over the intervention's candidate set; this convention is shared across the randomized interventions and prevents a zero pre-selection NormRouter score from turning the probe into a skipped layer.
For \methodlock we report only the five-task averages, 47.05 (\textsc{Top-K}) and 43.77 (\textsc{Random}) at 469M (Table~\ref{tab:own_model_randomization}); its per-task scores are not included in Table~\ref{tab:observation_own_models_full}.

\begin{table}[h]
\centering
\caption{Routing-randomization results on our trained models. Accuracy (\%) is reported on five downstream benchmarks. \textsc{Top-K}: learned top-$k$ routing; \textsc{Random}: mean accuracy after randomizing one deep-half MoE layer at a time and averaging across layers. For \methodfull, $^\dagger$ denotes the matched top-8 random protocol and unmarked \textsc{Random} denotes the standard full-pool random protocol. ARC-E, ARC-C, PIQA, and HellaSwag use length-normalized accuracy (\texttt{acc\_norm}) and WinoGrande uses \texttt{acc}, so Top-K values differ from the raw-accuracy values in Table~\ref{tab:downstream}. \methodlock is reported by its averages only (Table~\ref{tab:own_model_randomization}).}
\label{tab:observation_own_models_full}
\vspace{4pt}
\resizebox{\textwidth}{!}{%
\begin{tabular}{llcccccc}
\toprule
\textbf{Model} & \textbf{Routing} & \textbf{ARC-E} & \textbf{ARC-C} & \textbf{PIQA} & \textbf{HellaSwag} & \textbf{WinoGrande} & \textbf{Avg} \\
\midrule
\multirow{2}{*}{Vanilla MoE (469M)}
 & \textsc{Top-K}               & 44.70 & 25.09 & 65.94 & 38.63 & 51.14 & 45.10 \\
 & \textsc{Random}              & 43.05 & 24.82 & 63.43 & 37.73 & 50.12 & 43.83 ($-1.3$) \\
\midrule
\multirow{3}{*}{\methodfull (469M)}
 & \textsc{Top-K}               & 47.39 & 25.94 & 69.10 & 40.64 & 52.72 & 47.16 \\
 & \textsc{Random}              & 42.06 & 24.76 & 60.30 & 37.26 & 52.32 & 43.34 ($-3.8$) \\
 & \textsc{Random}$^\dagger$    & 41.61 & 25.21 & 60.62 & 37.46 & 50.61 & 43.10 ($-4.1$) \\
\bottomrule
\end{tabular}%
}
\end{table}

\section{Pool Auxiliary Loss: Detailed Derivation}
\label{app:pool_aux_derivation}

Here we provide the full derivation showing that the pool-level loss decomposes into per-layer terms.

Starting from the pool loss definition:
\begin{align}
    \mathcal{L}_{\text{pool}} &= \alpha_{\text{pool}} \cdot M \cdot \sum_{i=1}^{M} \overline{f}_i \cdot \overline{P}_i \\
    &= \alpha_{\text{pool}} \cdot M \cdot \sum_{i=1}^{M} \overline{f}_i \cdot \frac{1}{L}\sum_{l=1}^{L} P_i^{(l)} \\
    &= \frac{1}{L} \sum_{l=1}^{L} \alpha_{\text{pool}} \cdot M \cdot \sum_{i=1}^{M} \overline{f}_i \cdot P_i^{(l)}.
\end{align}

The last step uses the fact that $\overline{f}_i$ does not depend on $l$.
Each summand is the per-layer pool loss contribution, which can be computed independently.

\paragraph{One-step-behind computation.}
Computing $\overline{f}_i$ requires statistics from all layers, which are unavailable until the full forward pass completes.
To avoid cross-layer tensor dependencies (which would break activation checkpointing), we use a \emph{one-step-behind} scheme: each layer computes its pool loss contribution using $\overline{f}_i$ from the \emph{previous} micro-batch.
The global token distribution $\overline{f}_i$ is accumulated without gradients and updated after all layers complete their forward pass.
Only the routing probabilities $P_i^{(l)}$ carry gradients, so the pool loss only updates router parameters, not expert FFN parameters, through this path.

\section{NormRouter: Monte Carlo Initialization Details}
\label{app:monte_carlo}
The main text uses $c$ as a fixed calibration factor for the NormRouter score scale.
Given $E$ experts and top-$k$ routing, we choose $c$ so that the initial selected scores have approximately unit magnitude:
\begin{equation}
    c = \mathbb{E}\!\left[\frac{1}{\sqrt{\sum_{j=1}^{k} y_{(j)}^2}}\right], \quad
    y = \text{ReLU}\!\left(\frac{x}{\|x\|_2}\right), \; x \sim \mathcal{N}(0, I_E),
\end{equation}
where $y_{(j)}$ denotes the $j$-th largest component of $y$.
Algorithm~\ref{alg:monte_carlo} estimates this expectation by Monte Carlo sampling at initialization time.

\begin{algorithm}[h]
\caption{Monte Carlo estimation of NormRouter scale constant $c$}
\label{alg:monte_carlo}
\KwIn{Number of experts $E$, top-$k$ value $k$, number of samples $N = 10^5$}
\KwOut{Scale constant $c$}
$\text{samples} \leftarrow []$\;
\For{$n = 1$ \KwTo $N$}{
    Sample $x \sim \mathcal{N}(0, I_E)$\;
    $y \leftarrow \text{ReLU}(x / \|x\|_2)$\;
    Sort $y$ in descending order, take top-$k$ components $y_k$\;
    $\text{samples}.\text{append}(1 / \sqrt{\sum_j y_{k,j}^2})$\;
}
$c \leftarrow \text{mean}(\text{samples})$\;
\Return{$c$}
\end{algorithm}

\end{document}